\documentclass[10pt,twocolumn]{article}
\usepackage{graphics}
\usepackage{algorithm2e}
\pdfoutput=1
\begin{document}
\title{A Learning Algorithm based on High School Teaching Wisdom}
\author{Ninan Sajeeth Philip}
\thanks{N. S. Philip is with the department of Physics, St.Thomas College in Kozhencheri, India\\ e-mail: nspp@iucaa.ernet.in}
\maketitle

\begin{abstract}
A learning algorithm based on primary school teaching and learning is presented. The methodology is to continuously evaluate the performance of the network and to train it on the examples for which they repeatedly fail, until, all the examples are correctly classified. Empirical analysis on UCI data show that the algorithm  produces a good training data and improves the generalization ability of the network on unseen data. The algorithm has interesting applications in data mining, model evaluations and rare objects discovery.
\end{abstract}


\section{Introduction}
The efficiency of any machine learning algorithm is reflected in its generalization ability, which is defined as the ability of the algorithm to give the accuracy it produces on the training data to a set of similar, but unseen test data. However, generalization ability is known to depend explicitly on the quality of the training data \cite{Wann1990}.  Thus, selection of a good sample for training a network (hereafter used to mean any implementation of a machine learning algorithm) plays a crucial role in its evaluation. To this end, random sampling of the data and the use of validation data to estimate the quality of learning are very popular in the machine learning community. If the data size is large and uniform, random sampling is likely to pick up examples that are mostly seen in each class, from regions close to the mean of the Gaussian distributions representing the classes. The logic in such a selection is that, if a machine is trained on the mean of the distributions, it is very likely to do well on the rest of the data clustering around it. While this is acceptable in many cases, when there are sparse representation of examples within a class, it is very unlikely that such examples are collected by the random sampling process. This problem is addressed in the case of speech recognition using Turing's estimate in \cite{Katz87estimationof} where different levels of sampling based on the reestimate of the sparseness in a data is used to identify a good training data. While many algorithms for preparing  a good training data are found in literature, optimal data selection for training neural network continues to be an area of active research \cite{Keeni2002,Vijayakumar1998, Setiono97,Choueiki1999,Masashi2000, Kazuyuki2000, Wang2005, Cano06onthe, Carlos2006}.

Another popular approach to prepare a good (error free) training data is based on Wilson's editing scheme in which a k-NN classifier is used to identify outliers and incorrectly labeled examples in a training sample \cite{WilsonEditing}. In this scheme, a sample is retained in the training data (labeled as good) if and only if the k-NN classifier is able to correctly predict the class of the sample even after removing it from the training data. Wilson's editing is the basis of subsequent development of editing algorithms such as multi-edit \cite{MultiEdit}, citation editing \cite{CitationIndexing}, supervised clustering editing \cite{SCEditing} etc. In this paper, an alternate optimal data selection criteria is proposed, in which, we start with a null training set and instead of removing the failed entries (outliers) from the training sample, identify such objects from the test data and add them to the training data. In contrast, objects that pass the classifications are retained in the test sample. The outcome is just the opposite of Wilson's editing scheme. Here, instead of a clean training sample, the training is done on the outliers and the boundary examples in the data. To prevent the network from incorrectly learning on the outliers, sufficient ''good'' examples are added to the training data in such a way that the network correctly predicts the class of the object even when the outliers are present in the training data. The argument is based on information theory perspective that such a training sample will carry information relating to all possible variants in a real world situation.

The proposed scheme has much in common with primary schooling that is considered to be a decisive learning period in the educational stream of a student. Teachers at this stage give personal attention to students to compensate for their learning inabilities by repeatedly training them on their failures. Repeated training is believed to enable the student to gradually comprehend the underlying concepts so that learning can progress. Continued evaluation on the basis of questionnaire or dictations are used by the teacher to estimate the learning curve of individual students. This information is used to select topics that are to be emphasized in the next teaching cycle. In this paper, a similar procedure for the optimal selection of training examples for a machine learning algorithm is attempted.

It is understood that machine learning algorithms make their predictions on the basis of some flexible evaluation criteria, usually a function of the connection weights, that favor one of the possible outcomes for a given input pattern. The training process adjusts the criteria to reduce the overall error in the predictions of the network to a global minimum.  Although, the common practice is to ignore the marginal differences between these evaluations and to simply consider the one with maximum score as the possible outcome, it is possible to estimate each score values and to put it as a confidence measure on the predictions \cite{Wei1996}. For example, in Bayesian classifiers, every prediction is associated with a Bayesian posterior probability that gives a quantitative estimate of the confidence with which a prediction is made. This paper discusses the use of an Optimal training Data Selection Algorithm (ODSA) for a Difference Boosting Neural Network(DBNN) \cite{nsp2000}, in which the scheme can be readily implemented. The DBNN software source code is available at \cite{NSPHp}.

\section{The Difference Boosting Neural Network}
Bayesian Learning, though powerful, is computationally intensive due to the conditional dependence of parameters on the posterior computations. A naive approach is to assume conditional independence \cite{Dawid97conditionalindependence} of features so that the total probability may be computed as the product of individual probabilities. This is known as Naive Bayes classification. For a tutorial introduction to Naive Bayes see \cite{Moore, elkan97}. The major limitation of Naive Bayes is that it completely fails even on the typical XOR problem. So, one would like to ask whether it is possible to \textit{impose } conditional independence on a data without loss of accuracy in the posterior computation. The answer is positive and to explain how, let us consider the XOR problem were one has two independent input features, each of which can either be a 1 or a 0. Although the two inputs are independent variables, once the class is fixed, the value selected for any one feature imposes a constraint on the value the other feature can take. Thus, when class is a 1, if the first feature value is 1 the other feature can take only a value 0 and \textit{vice versa}. This may be represented as $$ 1 => [ \{1,0\};\{0,1\}]$$ Similarly when class is 0, $$ 0 => [\{1,1\}; \{0,0\}]$$ In other words, if we replace the independent input features of the XOR gate by a pair of new variables formed by combining all it's dependent variables, that new feature become conditionally independent and the overall posterior probability can thus be computed as the product of individual probabilities. Interestingly, when coming to non-binary real value features, the same criteria can be applied if we replace the binary values with the probability density distribution of the features.  The computation of the probability density can be simplified by binning the feature vector to give a histogram of the distribution. The histogram approximates the true distribution when the bin size is small. However the number of bins can't be arbitrarily chosen. Elementary statistics tells us that binning can approximate a distribution only if there are sufficient number of samples in each bin. Thus, in many practical situations, taking the bin size as $\sqrt{N}$ where N represents the number of discrete values taken by the variable is a good choice. Another possibility is to start with small number of bins and increase the number until the classification seems to reasonably separate the classes in the data. This is usually done when the features are continuous and $\sqrt{N}$ becomes large.

The implementation code for the above is straightforward. Initially one defines the bin size for each feature and constructs a memory model with count=1 in all the bins. Then each feature vector in the training sample is read and depending on the bin ($C_{in}$) into which they fall, the count in the bin is incremented. Each $C_{in}$ is also linked to a separate set of bins  ($C_{class}$) associated to the target classes in the data. These bins have a matrix structure, $C_{J,M}$, with the rows representing the other $J$ features and the column representing the histogram of their distributions in $M$ bins for the given $C_{in}$. Since the class is always known for the training sample, depending on the value of the features, the appropriate bins in the class are incremented. This is repeated until all the examples in the training data are read. We have now expressed our features into a set of conditionally independent partitions. It is thus possible to compute the likelihood for each class $C_k$ to produce a feature $U_m$ with a value in the bin \textit{m}, given by the intersection $${P}(U_{m}\cap C_{k})$$ as the ratio of the count in the bin $C_{U,m}$ to  $\sum_K C_{U,m}$, where $K$ is the total number of classes in the data. When there is no data in a bin, this defaults to a likelihood of $\frac{1}{k}$ as desired.

The Bayesian probability can now be computed if we also know the prior associated to each bin. The binning procedure makes it very difficult to define prior from the domain knowledge alone. If the distributions are distinct, prior can be taken as unity. But in many practical situations, there will be some overlap in the distribution of two or more classes.  The DBNN thus estimates the prior from the data during the training process by replacing it with a weight function that is updated as the training progress. It is given by $W_m$ in the Bayes theorem which now takes the form $$ P(C_{k}\mid U_m)=\frac{\hat{P}(U_{m}\cap C_{k}) W_{m}}{\sum_{K}\hat{P}(U_{m}\cap C_{K}) W_{m}}.
$$ Here $K$ represents the number of possible classes. Initially, a uniform small value ($\frac{1}{k}$) is assumed for $W_m$ and after each iteration, whenever a classification fails, it is incremented by a small value. This increment in weight for a incorrectly classified (\textit{failed}) example that has a feature value falling in bin \textit{m} is given by $$\Delta W_{m}=\alpha \left[ 1-\frac{P_{k}}{P^{*}_{k}}\right]  $$  Here  \( P_{k} \) represents the computed Bayesian probability for the actual class \( k \) and \( P_{k}^{*} \) that for the incorrectly represented class. Since it is computed only when \( P_{k}^{*}  > P_{k}\), it can be shown  that the value of $\Delta W_{m}$ is always positive and is bounded by the upper and lower limits $[ \alpha, 0]$. The number of iterations and the increment fraction $\alpha$ can be independently chosen. A typical value of 10 iterations and increment factor 0.9 (also known as learning parameter) are used in this paper. It is this boosting or weightage function proportional to the difference of the probabilities that gives the name \textit{difference boosting} to the network.

Since DBNN uses Bayesian probability as its central rule for decision making, the computed posterior probability gives the confidence the network has in its predictions. When the events are equally likely, the posterior will be same for all outcomes and hence will be low. In a two class problem, this means 50:50 and for a K class problem, this will be $\frac{1}{K}$. In the examples described in this paper, the posterior probability ranges between 0 and 100 and is  hereafter referred to as confidence level. The DBNN is widely used in astronomy research where one has to deal with large data with high uncertainty in their observed feature values \cite{steve,nsp,goderya,Rita2006}.
\section{The Importance of the Training Sample}\label{TrImp}
\begin{figure}
 \centering{\resizebox*{0.45\textwidth}{0.2\textheight}{\includegraphics{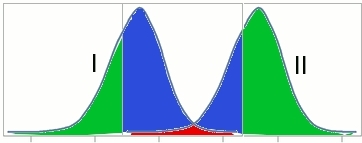}}}
 \caption{The distribution of two samples I and II are shown with their tails partially overlapping each other. The clean (non overlapping) examples in the sample are represented by the green region while the blue and red regions represent the samples belonging to two different classes with same feature values. Due to the dominance of objects shown as blue, it is very likely that all objects in this region are classified to the  class of blue objects, resulting in the wrong identification of the objects shown in red.}\label{fig:SampleDist}
\end{figure}
Bayesian classifiers have the ability to marginalise outliers and apply the Occam's razer principle by default. Then do we need an optimal training data? Why not train on the entire data and marginalise the outliers? This appears to be a straightforward solution and it works in the case of most outliers. But outliers are not the only issue here. What about sparsely represented examples? This is illustrated in figure \ref{fig:SampleDist}. As long as the distributions are distinct, it is possible to make correct predictions and we can call them clean examples. In the figure, the green regions represents the distribution of the clean samples in the data. They can be classified without any error with high confidence using Bayesian theorem that estimate the posterior probability as the product of the likelihood and the prior across the different classes. Likelihood, the observed frequency with which an event is related to the cause, is represented as $P(B\mid A)$ where $B$ is the observation and $A$ is the cause. In the figure, for the overlapping regions, likelihood for the examples in blue are higher than the likelihood of the examples shown as red. As a result, it may happen that the examples in red are incorrectly classified. If these are the only available observation for the objects, prior alone cannot compensate for the correct classification of the examples in the red region. Even when there are other observations (features), the large number of examples in the blue region of one class may dominate the Bayesian posterior estimation and may result in the incorrect classification of the few red examples. The only way to overcome this is to suppress the prominence of the blue examples in the likelihood estimation so that the differences in other features may become significant in the Bayesian classification. This means, if a selection criteria is defined such that a minimum number of examples from each class that are just required for the correct classification of the objects in its class are added into the training sample, that training may demonstrate better performance on the data than random samples or whole data used for training the classifier. An ideal training data for the example in Figure \ref{fig:SampleDist} is thus the one from which the blue examples are removed until the differences in other features may become significant in the classification. Although it might appear straightforward to do such a sampling, when it comes to multiple feature data, the addition of one sample can have multiple effects and the selection could become a nontrivial optimisation problem. In this context, it is argued that the primary school teaching method of going one step at a time, compensating for the complications in each iteration, will converge to an optimal training data in a finite number of rounds. This is what is attempted in the ODSA algorithm that is described next.

\section{ODSA Algorithm}
As stated in the previous section, our objective is to prepare a training data that gives a balanced likelihood estimate for all the classes so that even subtle differences in features can help the correct classification of the data. Let us consider a classification problem where one has to label an object into one of the $K$ possible predefined classes on the basis of some observed pattern of features. If all the outcomes are equally likely, the confidence in a prediction is $\frac{1}{K}$, which is generally called a flat prior. We start our estimates with such a flat prior and randomly pick up one data example from each class as our initial training sample to train the network. Since there are no confusing examples, the training accuracy on the training data can go close to 100\%. The trained network is then used to predict the class of the objects in the (test) data from which the training samples where taken. If the samples used for training are similar to that of the entire data, the test data should give a classification accuracy comparable to that of the training data. However, in practice, only a fraction of the test examples will be correctly predicted. At this stage, we search through the predicted samples to select one more example from each class to be added to the training data. The criteria for selection are that
\begin{enumerate}
 \item the example added to the training data is a failed example
\item it is  predicted with the maximum confidence into the wrong class
\end{enumerate}
\begin{figure}
 \centering{\resizebox*{0.45\textwidth}{0.3\textheight}{\includegraphics{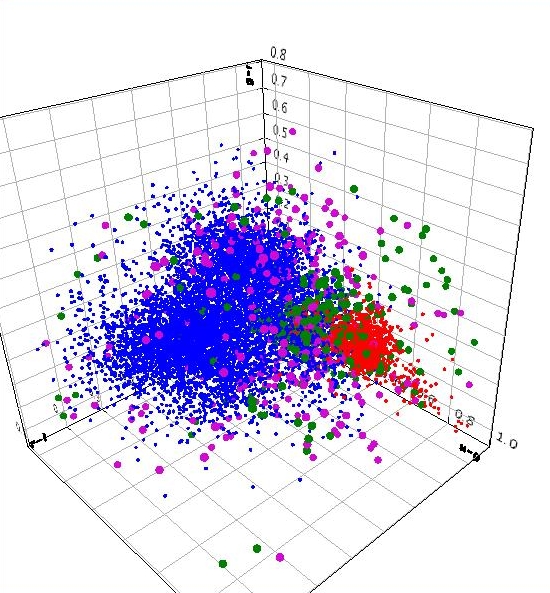}}}
 \caption{The projection of optimally selected training samples for a two class problem in feature space. The test data samples are shown in blue and red and the optimally selected representative training samples are shown as green and pink dots. It can be seen that the selected training examples are mostly from the boundaries as stated in text.}\label{fig:hypersurf}
\end{figure}
  The fact that it failed with maximum confidence means that it has features very similar to the examples in the other class (it is on the boundary of the decision surface). The network is now trained on the $2\times K$ samples and is tested on the remaining data as before to identify the failures. Since we now have one competing example for each of the incorrectly identified examples in the previous round, the likelihood for those examples will be halved in this round. So the failures in this round are likely to be caused by the other examples. With the same above criteria, one more failed sample from each class is added to the training data. In every step, a maximum of $K$ samples are removed from the test data and are added to the training data. In the third round, the alteration of the likelihood due to the addition of the examples in each class is bounded by a lower limit of $\frac{1}{3}$ and upper limit $\frac{1}{2}$. In round $n$, it will be bounded by lower limit of $\frac{1}{n}$ and upper limit $\frac{1}{2}$. This means, as the number of iterations increase, the fluctuations in the likelihood decreases asymptotically and the algorithm will eventually converge. The procedure is repeated until all the unseen samples (test samples) are correctly classified. It may be noted that the training data need not give 100\% classification accuracy any more. This is because, the process has removed most of the difficult, boundary examples and outliers from the test data and added them into the training data. The ODSA algorithm is given in Algorithm \ref{algo:ODSA}.

\begin{algorithm}
 \SetAlgoLined
 \KwData{Test Data (X) with $k$ classes and a null-set of training data (Y)}
 \KwResult{An optimally selected training data}

 initialization\;
     Remove one example for each class from test data and add them into the training data\;
     This gives a training sample of $K$ examples to start with.

   Train the classifier on the training data Y\;
   Test the classifier on the Test data X\;

   \While{There are errors in prediction }{
      start from beginning of results file\;
     \For{$i\leftarrow 1$ \KwTo Number of classes in data  ($K$) }{
      \While{not end of the test results file}{
        \If{This is the incorrect prediction with highest probability in this class?}{
             remove it from the test data X and add it to the training data Y\;

             }
             \Else{
             continue searching for the data that is incorrectly predicted with highest probability from this class\;
             }
      }
     }
   Train the classifier on the training data Y\;
   Test the classifier on the Test data X\;

  }

  Save the training and test data and exit\;

\caption{The ODSA Algorithm}\label{algo:ODSA}
\end{algorithm}

Picking up the example that fails with maximum confidence from the test data and adding it to the training sample is the crucial step here. In each iteration, the example that is added to the training data is a failed example that looks similar to an object in another class. This process will collect all the boundary examples from the test data to the training data. What about outliers and incorrect labels? When outliers and incorrectly labeled examples are added into the training data, that may cause ''good'' examples to be classified wrongly by the network. So the algorithm in its subsequent iterations will add the failed and good examples that looks similar to them to the training data. This is repeated until the likelihood for the class of the good examples eventually marginalise the effect of the outliers that are fewer than the good examples.  This is the major difference of the proposed method from Wilson's editing scheme. In the teacher-student scenario, this is how a teacher would correct a \textit{misunderstanding} that causes the student to come to wrong conclusions. A typical example\footnote{This figure was plotted with VOPlot software, courtesy: Rita Sinha} of the training samples selected in a two class problem where the decision surface is nonlinear is shown in Figure \ref{fig:hypersurf}.

An important point to be noted here is that training is done only on the optimally selected samples. The rest of the data is unseen by the network during its training epochs. In brief, by regulating the likelihood estimates through the sample selection method and by computing the prior by difference boosting, the algorithm identifies a set of examples that can be used to train the data much better than what would be possible through random selection or perhaps Wilson's editing algorithm where only good examples are used for training.
\section{The Learning Curve}
Just like the algorithm, the learning curve produced by the network also has similarity to human learning curves. In most practical problems, the boundaries of the classes would be fuzzy and it may become difficult even for human experts to confidently classify examples in those regions. This need not be due to the vagueness in the class definitions; rather, it could be due to the feature estimation inaccuracies in some regions of the feature space. For example, consider the automated classification of deep sky images by a sky survey project. Unlike the set conditions in a laboratory, the astronomer, in this case has to infer and classify objects on the basis of their appearance through his telescopes, spectrograph etc and that introduces feature estimation inaccuracies.

\begin{figure}
 \centering{\resizebox*{0.49\textwidth}{0.27\textheight}{\includegraphics{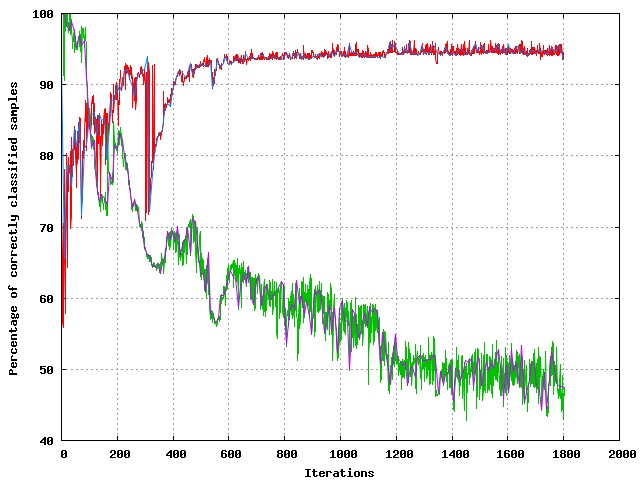}}}
 \caption{The learning curves in a typical classification problem with a high degree of uncertainty and probable incorrect labeling in the training sample is shown. The Pseudo-learning curve is shown as green and the real-learning curve is shown as red. A smooth curve is over plotted to show the trends at each epoch of learning. Note the spikes in the learning process. In this example there are about a million objects in the test sample and a few thousand objects in the training sample at the last iteration shown. }\label{fig:learningcrv}
\end{figure}

For convenience, we divide learning curves into two groups, namely (a) Pseudo-learning curve (b) Real-learning curve. The Pseudo-learning curve is the learning accuracy obtained on the training sample and Real is the same on the test sample. Initially, when the training examples are only a few, the pseudo-learning level is close to 100\% and the real-learning level is much lower. Gradually, when more and more failed (boundary) examples get added to the training sample, the classification accuracy for the training data start to decline and that for the test data start to climb. Figure \ref{fig:learningcrv} illustrates the learning curve in the case of a data with a high degree of uncertainty in labeling as well as in the estimation of feature parameters. This is typical of many practical situations in space research, the details of which are not relevant for the current discussion.

It may be noted from the figure that initially, with increase in iterations, there is a rapid decline in the prediction accuracy of the training sample (represented as green) and a similar increase in the prediction accuracy of the test sample (represented by red lines). This is the time at which the major boundaries of the classes are identified by the network and that accounts for the improved prediction accuracies on the test data. As the learning progresses, sudden spikes appear in the learning curve of the test samples causing their prediction accuracies to fall intermittently. This is mostly caused by the addition of incorrectly labeled objects to the training data. Since the algorithm picks up objects that have failed with maximum confidence, it turns out that all the incorrectly labeled objects are picked up once the class boundaries get defined. However, as long as such incorrect labeling is not intentional and are only a few in number, the likelihood for the correct examples increases and the Bayesian algorithm will start ignoring them so that the network regains its classification accuracy in subsequent iterations. The training accuracy represented by the Pseudo-learning curve will however continue to remain low for some more time as almost all the incorrect labels are now getting gathered by it. In other words, this low level is not due to the inefficiency of the network algorithm, rather is caused by the incorrect labels assigned to them in the dataset, against which the prediction results are compared. This is substantiated by the increasing accuracy in the test data which is much larger than the training data.

Although the test accuracy continues to rise very close to about 100\%, it may be noted that spikes reappear for a second time causing degradation in accuracy on the test sample. Examining the data that were selected at these times showed that this was mostly due to the network's attempt to redefine the boundary of the classes based on the updated training data. As the iterations continue, the network will eventually predict all the samples in the test data correctly and will at the same time have all the boundary examples and false label examples collected in its training sample. As a result, the accuracy for the optimal training data will be less than or equal to that of the test data. To get a more realistic estimate of accuracy, after training on the selected samples, the prediction accuracy on the entire data is reported as the network accuracy.
\section{Illustration of the advantage of the scheme}

Here, Landsat Satellite data from UCI repository is used to illustrate the advantage of the scheme. This database consists of the multi-spectral values of pixels in 3x3 neighborhoods in a satellite image and the classification associated with the central pixel in each neighborhood. The aim is to predict this classification, given the multi-spectral values. In the sample database, the class of a pixel is coded as a number and are represented by numbers 1 to 7  respectively referring to:
\begin{enumerate}
 \item  red soil
\item  cotton crop
\item grey soil
\item damp grey soil
\item soil with vegetation stubble
\item mixture class (all types present)
\item very damp grey soil
\end{enumerate}
There are no examples with class 6 in this dataset. The data available at the UCI repository is split into a random training sample of about 4000 examples and a test sample of $\sim$ 2000 examples for uniformity in evaluations. The best reported accuracy on this dataset is 92\% \cite{DTREG}.

\begin{table*}
\begin{center}
\begin{tabular}{lll}
Data Name & Sample Size & Accuracy \\
\hline
Original Training Sample & 4435 & 95.3\% \\
Original Test Sample & 2000 & 84.1\% \\
Optimally Selected Training Sample & 1063 & 98.77\%\\
Original Test Sample & 2000 &85.35\%\\
\end{tabular}\caption{The optimal data selection algorithm was performed on the original training sample of Landsat Satellite data from the UCI repository. The table shows the prediction accuracies obtained on the unseen test sample with the original training data and with the optimally selected subset. Note the improved training and test accuracies with the optimally selected training subset. This is because optimal selection has compensated for some of the outliers that usually have adverse effect on network performance.}\label{tab:OSA1}
\end{center}
\end{table*}

\begin{table*}
\begin{center}
\begin{tabular}{lll}
Data Name & Sample Size & Accuracy \\
\hline
Optimally Selected Training Sample & 1478 & 98.65\%\\
Entire data & 6435 &99.64 \%\\
Original Training Sample & 4435 & 99.75\% \\
Original Test Sample & 2000 & 99.4\% \\
\end{tabular}\caption{The optimal data selection algorithm was performed on the combined training and test samples of Landsat Satellite data from UCI repository. The table shows the prediction accuracies obtained on the entire data, original test data and the original training data sets. There is a remarkable improvement in the overall prediction accuracies and this is by about 7\% higher than the accuracy ever reported on this dataset.}\label{tab:OSA2}
\end{center}
\end{table*}

We will use this data to show how an optimal data selection algorithm can be used to generate a training data that has maximum information density per pattern. In the first example we train a DBNN on the supplied training data and test it on the test dataset. Then we do an optimal data selection from the same training sample (ie 2000 examples) and train DBNN on it. The testing is done on the same test dataset provided by the UCI repository. If the optimally selected data contains all the information in the original training data, we should expect a comparable accuracy on the test data in both the test rounds. The result in Table \ref{tab:OSA1} shows that the accuracies in both cases are comparable on the unseen test sample, confirming that all the information contained in the original training data is now packed in the smaller subset of it generated by the data selection algorithm. The accuracy obtained on the original training sample is comparable to the values reported in literature for Bayesian and Belief networks \cite{Belief2006,Bayes1997}. In fact, there is a slight improvement in the test accuracy with ODSA due to the controlled addition of samples to the training data as explained section \ref{TrImp}.

In the second example, we mix both the training and test datasets from UCI and then run optimal data selection algorithm on it. This should produce a training data, again, with maximum information density per pattern. The quality of the algorithm is then tested against the entire dataset and the individual training and test datasets in the UCI repository. The results are shown in Table \ref{tab:OSA2}. It is found that a remarkable improvement by about 7\% over the best result on this dataset in literature is attained by the network.

A similar analysis was done on a few more datasets from the UCI repository and the best known prediction accuracies for those dataset in literature along with the their new prediction accuracies and training sample sizes are given in Table \ref{tab:othersC}.

\begin{table*}
 \begin{center}
\begin{tabular}{llllll}

Data & Original   & Best Test & ODSA & ODSA & Test \\
Name & $Tr_d$ Size & Accuracy & $Tr_d$ Size & Accuracy & data size\\
\hline
Shuttle & 43406 & \textbf{100\%} & 192 & 99.92\% & 14500\\
Heart & Random & 87\% & 131 & \textbf{100\%} &270\\
German & Random & 80\% & 528 & \textbf{93.2\%} &1000\\
Iris & Random & 98\% & 34 & \textbf{100\%} & 150\\
glass & Random & 88\%& 79 & \textbf{100\%} & 214\\
W.Breast Cancer& Random &96\% &70 &\textbf{100\%}& 629\\
Ecoli & Random & 88\% & 146 &\textbf{97.92\%}& 336\\
Haberman & Random & 76\% & 131 & \textbf{91.83\%}& 306\\
Ionosphere & Random & 95\% & 66 &\textbf{100\%}& 351
 \end{tabular}\caption{Best classification accuracy reported in literature (Col.3, as given in UCI repository) on a few datasets from UCI repository and the accuracy obtained with a DBNN trained on the optimally selected training data (Col.5) are shown. In cases where UCI provided a training data, ODSA algorithm was run on it. Otherwise, the entire data was used. In either case, training was done only on the optimally selected sample and the rest of the data was unused in the training epochs.}\label{tab:othersC}
 \end{center}

\end{table*}

Other than the improved over all classification accuracy, what could be more interesting here are a few failed samples. This is mostly the subject of the rest of the paper.
\section{Information from Failures}

While classification and catalog preparation could be one application of using a machine learning tool, a much more interesting and useful application is the study of the outliers or the objects that get labeled with relatively lower confidence levels. As mentioned before, Bayesian rule gives a very precise measure of the confidence (posterior probability) in every prediction. This confidence level indicates how well the class of the test object is defined by the given evidences (observations). We will briefly discuss how low confidence levels may be used to discover the inherent limitations in a system by taking an example from astronomy.

\subsection{Instrumental Limitations}

Quasars are deep sky objects that look like stars, but  in reality are extremely active galactic centers flashing out intense amounts of radiation that out shines the rest of the galaxy. Although they were first discovered in the radio-band of wavelengths, it is now known that there are many radio quiet quasars as well. Astronomers believe that these objects have a very massive black hole in its center and is swallowing surrounding matter that is gravitationally pulled to it at speeds close to that of light, resulting in the emission of intense radiation. The Doppler effect due to these relativistic velocities can be seen in the spectra of quasars and this is the  only confirmatory test for their identity.

The limitation with spectroscopy is that, since the objects are very distant and therefore faint, the integration time to obtain a reasonably good spectrum from these objects is prohibitably large even with modern large telescopes and highly sensitive devices. As for example, the Sloan Digital Sky Survey (SDSS) which is one of the most extensive surveys that discovered the highest number of quasars has only less than 1\% of its archived images with a spectrum.

A simpler alternative to obtaining the spectrum is to image the objects through different band-pass filters and then to consider the relative flux in each filter, which astronomers call the colour of the object. Quasar colours, like their spectra, are different from those of other objects. Since there could be many filters, as many colours also can be produced for each object. For example, SDSS has five filters and hence can produce four independent colour images for each object. It is thus possible to input these colours to a neural network and train it with spectroscopically identified objects, so that, later, when only the colour information is given as input, the network will correctly predict the identity of the object without the need of a spectra.

Although the procedure appears to be straightforward, quasars have overlapping colours with other star like objects. Also, spectral redshift due to the expansion of the universe introduces a drift in the colours of the quasars and degrades the classification efficiency at regions where the shifted colours becomes identical to the apparent colours of other types of objects in the sky. This causes the ODSA algorithm to add many samples from such overlapping regions to the training data to minimize the overall error in classification. The increased number of similar examples reduces the likelihood for these objects and during the training epoch, the Bayesian network assigns only a lower confidence value to objects belonging to such overlapping patches in the feature space. Identification of such regions tells the observer how such limitations can be overcome by improving the observational parameters.
\begin{figure*}
 \centering{\resizebox*{0.5\textwidth}{0.3\textheight}{\includegraphics{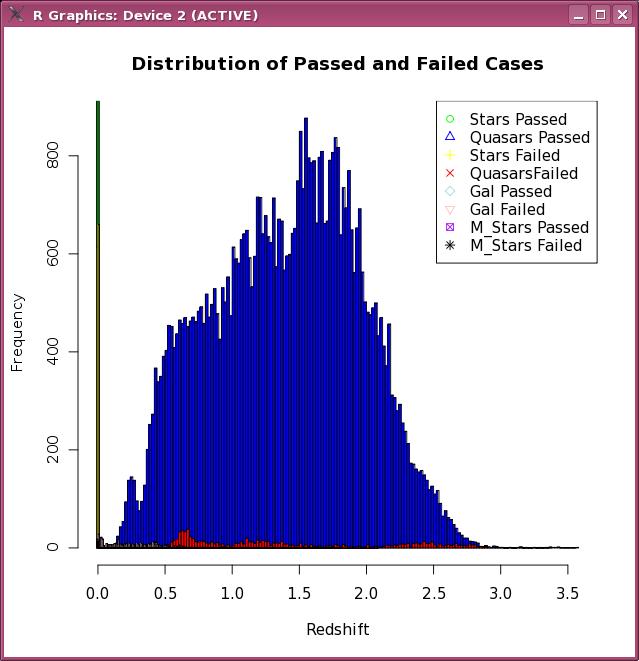}}\resizebox*{0.5\textwidth}{0.3\textheight}{\includegraphics{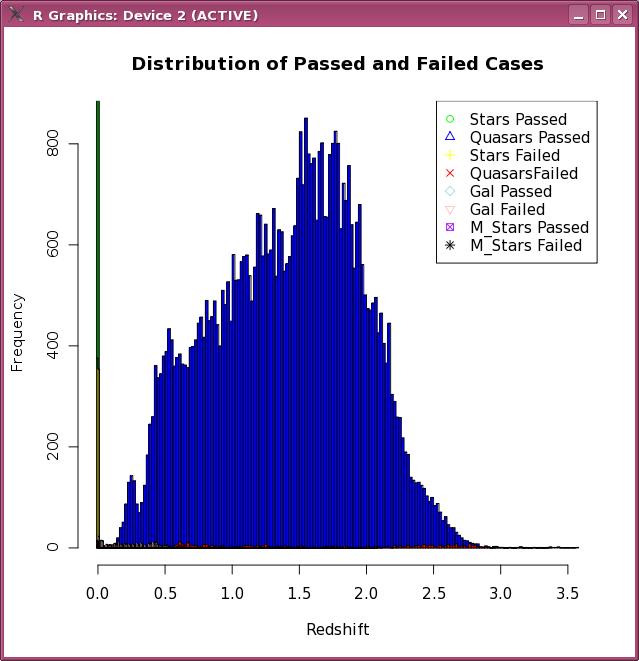}}}
 \caption{The LHS plot shows the failed objects in red and the RHS plot shows the histogram after removing the low confidence predictions. Since most objects that fail have low confidence, they are fewer in the RHS figure. The plots also reveal the limitation of SDSS colours and the redshift ranges at which such error patches appear. This information may be used to design newer filters that can produce additional colours to compensate for the instrumental limitations. }\label{fig:specqso}
\end{figure*}

Figure \ref{fig:specqso} shows the histogram of deep sky images classified from the SDSS DR6 data \cite{SDSS} release as a function of their redshift using the methodology described \cite{Rita2006}. The region selected was known to have a high concentration of quasars. In the plot to the left, blue are the correctly identified quasars and red are quasars that were incorrectly classified by the network. Stars, since they are from our own galaxy, have a redshift zero and hence they appear as a single line in the histogram. In the plot to the right, the histogram was plotted after removing all objects that had a confidence less than or equal to 56\%. This removed almost all the failed examples from the plot. However, it may be noted that it was not just the failed examples alone, but, a good number of correctly classified objects also got removed in the process. This is the cost of the confidence cut off. The important point to be noted here is that
\begin{itemize}
 \item  A detailed analysis of the features of those low confidence objects can be used to design and use additional filters that can resolve the colour space in those regions.
\item Outliers are not always objects with incorrect labels. Low confidence also could mean the presence of a totally different class of objects in the dataset. Since we have identified a sample of such objects, it is much more easier now to identify new classes of objects in the data, if they exists. In fact, quasars themselves were first discovered while investigating the outliers in colour plots.
\end{itemize}

\subsection{Model Evaluations}

Another practical application of the method is in the evaluation of models. Models can predict observationally \textit{falsifiable} results. One great example for this is the verification of the bending of light that glances the surface of the sun as a proof for the correctness of the model of gravity given by Einstein's general relativity. We address here a different problem to illustrate the application of machine learning tools to evaluate theoretical models in the light of observations.

Model evaluations are trivial where they predict precise values that can be verified with sufficient accuracy; as in the case of the bending of a ray of light in the presence of a massive object like the sun. However, in many areas of science such precise measurements are either not possible or are hidden from the observer. For example, the night sky has thousands of stars that appear in different colours and magnitudes across the horizon. Are they all of the same kind?

Astronomers who have studied these stars in detail have broadly classified them into
seven classes (O,B,A,F,G,K and M).
Since we are not interested in their classification schemes, we call them as 1,2,3 etc
instead of their astronomical names. Now, there are theoretical models that explain why
each star is distinct. The task is to check the accuracy of a given model on the basis of
available observations.
This is a nontrivial problem. Each star we observe is unique in its own way. The only
realistic measurement that can unambiguously identify the class of these object are their spectra. Since stars
are part of our galaxy, we can assume that the effect of redshift can be ignored in their
case. Even then, there are intragalatic dust and stellar reminiscence that can distort the
observed spectra that makes simple comparisons impractical.

The learning algorithm described above comes as a very useful tool in such situations.
The method is straightforward. In the stated example of stellar classification, use the
 theoretical model to predict all possible variants of the spectral types of the stars.
As mentioned above, each of these spectra will belong to one of the seven possible classes
which are known a-priori from the model.

Once the data is made, the evaluation round can be summarized as follows:

\begin{enumerate}

\item Use the algorithm to pick up the optimal number of training examples from the
data so that all the remaining samples are correctly classified by the network.
\item Test the network on the entire set of synthetic data and prepare a distribution
table for each class.

\item Test the network on real data and prepare a separate similar table.

\item If the model is good, the distributions in both tables should be comparable. If on the
other hand, the distributions are distinct in both the tables, the model cannot be accepted
without further evidence.
\end{enumerate}

This is illustrated below with an example in which 6300 samples \cite{Archana} generated
from 6 different stellar models were tested against 229 spectra of real
stars\footnote{The data is from Archana Bora and Ranjan Gupta,
Inter University Centre for Astronomy and Astrophysics; published with their permission.}.
Of these, 6199 good model spectra were used for our study. The network was trained on a subset of 682 examples that  was found to be sufficient to correctly classify \textit{all} the remaining data. The result of testing the network on the entire model
spectral data is shown in table \ref{tab:modelCM}. The last line of the table gives
the actual number of objects in each model (1,2,3 etc) as per the data. The last
column gives the total count in each model as per the network. In an ideal case,
both these numbers should be the same (as in the case of model 1). In reality,
this is very unlikely due to intrinsic differences within the same class of objects due to factors like dust and other observational limitations affecting the data. However, it may be noted that on this model data, most of the
objects (above 99 \%) were correctly identified.

\begin{table*}
\begin{center}
\begin{tabular}{|l|llllll|l|}
\hline
Label&1 & 2 & 3 & 4 & 5 & 6 & Total\\
\hline
1 &  1050 &   0   & 0  & 0 & 0 & 0 &  1050\\
2 & 0 & 1049&   0 & 0 & 0 & 0 &  1049\\
3 &0&  1 &1046 & 0 & 0 & 0 & 1047\\
4 &0 & 0 & 4 &1032 & 0 & 0 & 1036\\
5 &0 & 0 & 0 & 13 & 996 & 4 & 1013\\
6 &0 & 0 & 0 & 0  &  0  & 1004 &   1004\\
\hline
Total &1050 & 1050 & 1050 & 1045 & 996 & 1008 & 6199\\
\hline
\end{tabular}
\end{center}
\caption{The confusion matrix showing the class labels as per data (horizontal labels) and as per the network (vertical labels) for model data.}\label{tab:modelCM}
\begin{center}
\begin{tabular}{|l|llllll|l|}
\hline
Label&1 & 2 & 3 & 4 & 5 & 6 & Total\\
\hline
1&  6  &  9  &  0  &  0  &  0  &  0  &  15\\
2& 36  & 103 &  4 &   0 &   0  &  0  & 143\\
3&  0  &  2  & 43 &  16  &  0  &  0  &  61\\
4&  0  &  1  &  1 &   4  &  3  &  1  &  10\\
5&  0  &  0  &  0 &   0  &  0  &  0  &  0\\
6&  0  &  0  &  0 &   0  &  0  &  0  &  0\\
\hline
Total&  42  &  115  & 48 &  20  &  3  &   1  &     229\\
\hline
\end{tabular}
\end{center}
\caption{The confusion matrix showing the class labels as per data (horizontal labels) and as per the network (vertical labels) for real data. The reduced accuracy obtained in contrast to the model spectra is indicative of missing features in the model. }\label{tab:realCM}
\end{table*}

We then test the network on the real data. If the model is representative
of reality, the distribution of objects in this case should be comparable to that
in table \ref{tab:modelCM}. The confusion matrix for the real data is shown in
table \ref{tab:realCM}. It is seen that the network could not produce the same
accuracy on the samples for at least classes 1 and 4. Although classes 5 and 6
were also wrongly classified, the number of examples in those cases are too small
to validate the model.

The result in table \ref{tab:realCM} illustrates the contradictions
between theory and  observations. Given that the test sample size is very small,
this discrepancy could be argued to be just a matter of chance. On the contrary,
if this is the only observation we have, then there are two possibilities (1)
The model is inaccurate. (2) The feature representation is not good enough. Both
these issues are nontrivial problems and one cannot answer which is more likely
unless detailed investigations are done on the astronomy part. However, a simple
rule of thumb that machine learning people use reduces the likelihood for
option 2 to be correct. The logic is very simple. If the features used to represent
the sample were inefficient, one would not have obtained a good classification on
the model spectral data. The situation is just the opposite here. The network
correctly classified all the examples in class 1 of the model data while it
failed in the case of real data. This  demands a closer look on the goodness of the model.

A very pertinent question here obviously would be \textit{related to the
extent to which this information may be used to update a given model.} To answer this,
let us have a second look at the problem. We have seen that the network performed
reasonably well on the model spectra. This means that the models were distinct. The
failures occurred when we replaced the model spectra with actual spectra. A wrong
classification means that the actual spectra looked similar
to the model spectra from a different class to which it was incorrectly assigned by the network.
In other words, although the models were distinct, the modeling failed to comprehend
some distinctive features in the actual spectra. Since we now have the examples
that have failed and the models to which they got incorrectly identified with, analysis of
the averaged residual spectra obtained by subtracting the expected and obtained model
spectra from the actual spectra of each failed example should, in principle, highlight
the wrong priorities and missing features in the models. This information can be
used to update the theoretical model of the object until it is found that the accuracy obtained on real spectra is comparable to the accuracy obtained on the spectra created by the model.



\section{Conclusion}

The paper details an optimal data selection algorithm for selecting a complete
training sample for any neural network. By following a high school teaching pattern,
it is shown that the algorithm will be able to select the most valid information hidden
in a dataset. Some interesting applications of the network in discovery science and
model evaluations are discussed.

\section{Acknowledgments}
The author wish to thank the editor and the referees for their positive comments and suggestions to improve the manuscript. Thanks are due to the Inter University Centre for Astronomy and Astrophysics, Pune for providing their computing facility, Professors Ajit Kembhavi and Ranjan Gupta for their support in the preparation of the manuscript and the respective authors for providing the data used in this paper. This work was funded by the Indian Space Research Organization (ISRO) under RESPOND grant No: ISRO/RES/2/339/2007-08
\bibliographystyle{unsrt}
\bibliography{MyRef.bib}
\end{document}